\pdfoutput=1
\documentclass[11pt]{article}
\usepackage[preprint]{acl}
\usepackage{times}
\usepackage{latexsym}
\usepackage[T1]{fontenc}
\usepackage[utf8]{inputenc}
\usepackage{microtype}
\usepackage{inconsolata}
\usepackage{graphicx}
\usepackage{multirow}
\usepackage{booktabs}
\usepackage{amsmath}
\usepackage{algorithm}
\usepackage{algorithmic}
\usepackage{amsfonts}
\usepackage{subcaption}

\usepackage{listings}
\usepackage{enumitem}
\usepackage{soul}
\definecolor{myyellow}{rgb}{1, 0.972549, 0.584313}
\definecolor{mygreen}{rgb}{0.772549, 0.945098, 0.756862}
\definecolor{myblue}{rgb}{0.811764, 0.866666, 0.996078}
\definecolor{myred}{rgb}{0.984313, 0.749019, 0.737254}
\definecolor{darkgreen}{rgb}{0, 0.5001960, 0}
\definecolor{darkred}{rgb}{0.8, 0, 0}

\usepackage{xcolor}
\usepackage{colortbl}
\definecolor{marron}{HTML}{CC88B0}

\newcommand\blfootnote[1]{%
  \begingroup
  \renewcommand\thefootnote{}\footnote{#1}%
  \addtocounter{footnote}{-1}%
  \endgroup
}

\title{Scaling Data Difficulty: Improving Coding Models \\ via Reinforcement Learning on Fresh and Challenging Problems}

\author{
  Zongqian Li\textsuperscript{1, 2} \quad
  Tengchao Lv\textsuperscript{1} \quad
  Shaohan Huang\textsuperscript{1}\textsuperscript{†} \quad
  Yixuan Su\textsuperscript{1} \quad
  Qinzheng Sun\textsuperscript{1} \\
  \textbf{Qiufeng Yin\textsuperscript{1} \quad
  Ying Xin\textsuperscript{1} \quad
  Scarlett Li\textsuperscript{1} \quad
  Lei Cui\textsuperscript{1} \quad
  Nigel Collier\textsuperscript{2}\textsuperscript{†} \quad
  Furu Wei\textsuperscript{1}}\textsuperscript{†} \\[6pt]
  \textsuperscript{1}Microsoft Research \quad
  \textsuperscript{2}University of Cambridge
}

\begin{document}
\maketitle
\begin{abstract}
Training next-generation code generation models requires high-quality datasets, yet existing datasets face difficulty imbalance, format inconsistency, and data quality problems. We address these challenges through systematic data processing and difficulty scaling. We introduce a four-stage \textbf{Data Processing Framework} encompassing collection, processing, filtering, and verification, incorporating \textbf{Automatic Difficulty Filtering} via an LLM-based predict-calibrate-select framework that leverages multi-dimensional difficulty metrics across five weighted dimensions to retain challenging problems while removing simplistic ones. The resulting \textbf{MicroCoder dataset} comprises tens of thousands of curated real competitive programming problems from diverse platforms, emphasizing recency and difficulty. Evaluations on strictly unseen LiveCodeBench demonstrate that MicroCoder achieves 3x larger performance gains within 300 training steps compared to widely-used baseline datasets of comparable size, with consistent advantages under both GRPO and its variant training algorithms. The MicroCoder dataset delivers obvious improvements on medium and hard problems across different model sizes, achieving up to 17.2\% relative gains in overall performance where model capabilities are most stretched. These results validate that difficulty-aware data curation improves model performance on challenging tasks, providing multiple insights for dataset creation in code generation.
\end{abstract}

\blfootnote{\hspace*{-1.8em}\textsuperscript{†}Corresponding Authors. Project page: \\ \url{https://github.com/ZongqianLi/MicroCoder} }

\section{Introduction}
\label{Introduction}

\subsection{Background \& Related Work}

\textbf{Code Generation Datasets.} Training high-performance code generation models requires diverse, high-quality datasets. Early benchmarks like HumanEval \citep{chen2021evaluatinglargelanguagemodels} and MBPP \citep{austin2021programsynthesislargelanguage} established basic evaluation standards but were limited in scale. Human-curated competitive programming datasets emerged to address this gap: APPS \citep{hendrycks2021measuring} collected 10K problems, CodeContests \citep{Li_2022} provided 13K challenges, and TACO \citep{li2023tacotopicsalgorithmiccode} collected 26K problems from multiple platforms, all with verified solutions and comprehensive test cases. However, manual curation disadvantages led to the development of data generation. Code Alpaca \citep{codealpaca} and Evol Instruct \citep{luo2024wizardcoder} developed instruction-following datasets through LLM-based generation, while OSS Instruct \citep{wei2024magicoder} and Package Instruct \citep{huang-etal-2025-opencoder} increased coverage to 75K-110K problems by collecting open-source repositories. Recent work like KodCode \citep{xu-etal-2025-kodcode} introduced systematic generation pipelines combining question generation, self-verification, and chain-of-thought responses, achieving 447K verified problem-solution-test triplets. Despite these advantages, generated datasets often face diversity limitations, difficulty imbalances, and verification challenges.

\textbf{Data Quality and Difficulty Assessment.} Ensuring dataset quality requires addressing multiple dimensions: solution correctness, test case validity, and appropriate difficulty distributions. Web-collected data introduces noise including incomplete problem descriptions, incomplete test cases, and irrelevant content, necessitating robust filtering mechanisms. Self-verification approaches have emerged as a solution: OpenCoder \citep{huang-etal-2025-opencoder} employs teacher models to generate and execute test cases, while EvalPlus \citep{evalplus} combines LLM-generated tests with variation-based generation. The self-verification pipeline of KodCode 
validates correctness through unit test execution, though challenging problems that fail initial verification are typically removed. Difficulty assessment remains challenging, with most datasets relying on platform-specific labels that may not reflect model capabilities. Recent work uses model performance metrics (pass@k rates) as difficulty measures, while AceCoder \citep{zeng-etal-2025-acecoder} introduces automatic test-case generation for verification. These developments highlight an ongoing need for systematic approaches to data processing, quality assurance, and difficulty-aware dataset creation.

\subsection{Motivation}

Existing coding datasets face three limitations that influence effective model training:

\vspace{-6pt}

\begin{itemize}[left=0pt, itemsep=0pt, parsep=0pt]

\item \textbf{Difficulty and Recency:} Current datasets exhibit difficulty imbalance with simple problems dominating, lacking the challenging problems that drive model improvement. Additionally, most datasets lack recent problems, which are inherently harder as models have less pretraining familiarity with them.
\item \textbf{Format Consistency:} Training corpora mix diverse problem formats (function completion versus input/output) without standardized instructions, causing models to produce algorithmically correct solutions in incorrect execution formats.
\item \textbf{Data Quality:} Web-collected problems contain noise including incomplete descriptions, incomplete test cases, and irrelevant content, while test case distributions range from absent to excessively redundant.

\end{itemize}

\subsection{Contributions}
This work addresses the identified limitations through systematic data processing and difficulty-aware curation:
\vspace{-6pt}
\begin{itemize}[left=0pt, itemsep=0pt, parsep=0pt]
\item \textbf{Data Processing Framework:} We introduce a four-stage pipeline encompassing collection, processing, filtering, and verification, systematically addressing format inconsistency, data quality problems, problem difficulty and recency, and train-test overlap.
\item \textbf{Automatic Difficulty Filtering:} We propose a predict-calibrate-select approach leveraging LLM-based multi-dimensional difficulty metrics across five weighted dimensions to assess problem complexity, enabling removal of simplistic problems while retaining challenging problems that improve model generalization on difficult problems.
\item \textbf{MicroCoder Dataset:} We create a dataset of 13,300 real competitive programming problems from diverse platforms, emphasizing recency and difficulty, achieving 3× larger performance gains within 300 training steps compared to widely-used baseline datasets of comparable size.
\item \textbf{Key Insights:} We provide multiple insights for the relationship between problem difficulty and model generalization, the effectiveness of difficulty-based filtering across problem categories, and the combination of data curation strategies and training algorithms, demonstrating that difficulty-aware data curation achieves performance gains on challenging problems.
\end{itemize}

\begin{figure*}[t!]
    \centering
    \includegraphics[width=0.97\textwidth]{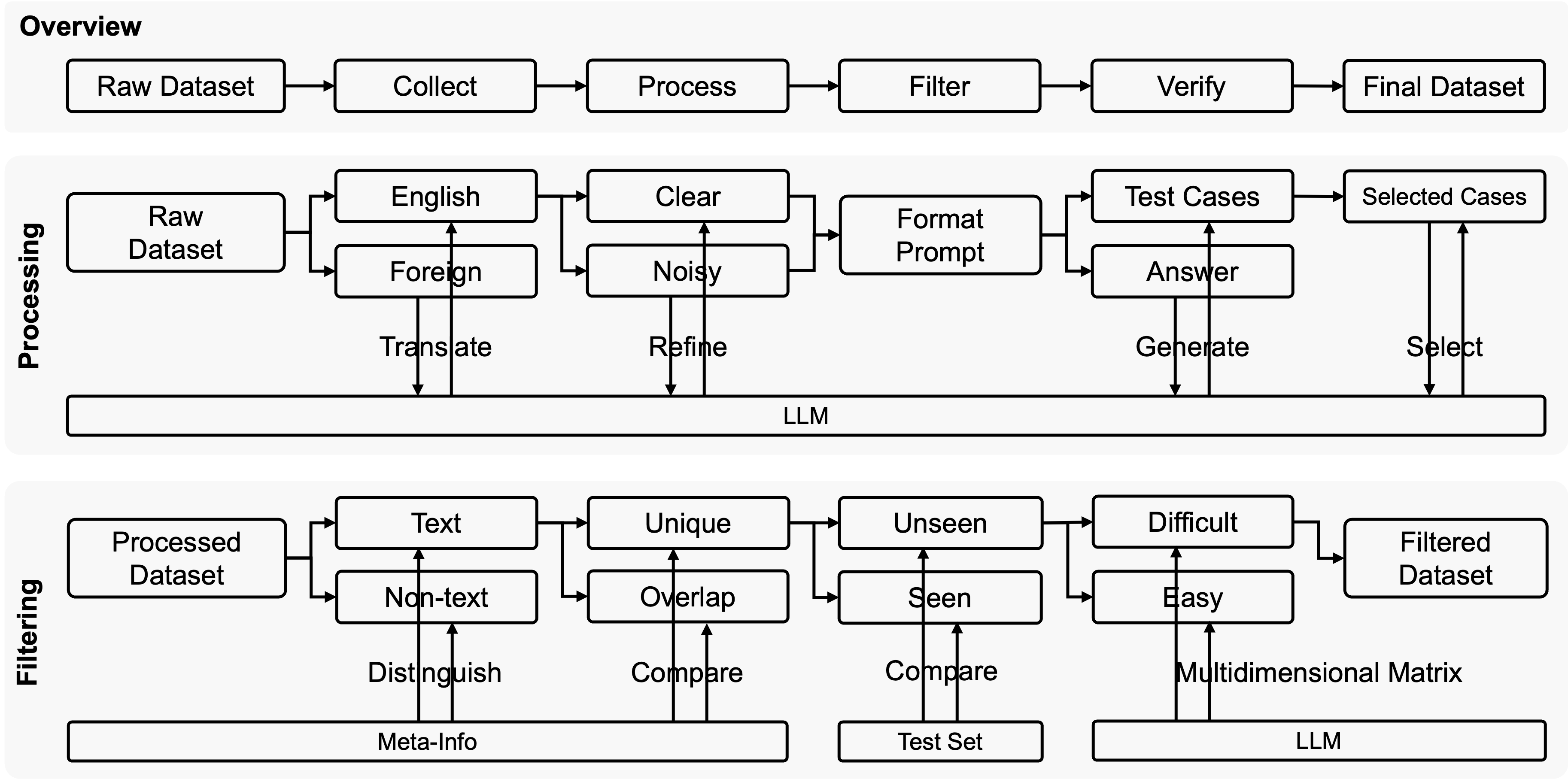}
    \caption{\textbf{Data Processing Framework Overview.} \textit{Top:} Four-stage pipeline transforming raw datasets into high-quality training data through data collection, standardization processing, quality filtering, correctness verification, and final dataset creation. \textit{Middle:} Standardization steps including translation of foreign language problems to English, noise removal and refinement, format prompt unification, and LLM-based test case generation and selection. \textit{Bottom:} Multi-stage filtering implementing hard requirements (text-only problems via content-type distinction, uniqueness via overlap identification, unseen status via test set comparison) and adaptive selection (difficulty-based filtering using multidimensional difficulty matrix with LLM to distinguish easy and difficult problems), producing the filtered dataset for verification.}
    \label{method}
    \label{Data Processing Framework Overview}
\end{figure*}

\section{Methods}
\label{Methods}

\subsection{Data Processing Framework}
\label{Data Processing Framework}

\subsubsection{Overview}
\label{Overview}

Our data processing framework (Figure \ref{Data Processing Framework Overview}) employs a four-stage pipeline to create high-quality coding datasets. The Collect stage collects data from diverse sources, including public datasets (Taco, KodCode, DeepCoder, etc.) and web-collected programming problems to maximize coverage. The Process stage standardizes data through language translation, noise removal, format normalization, and completeness validation. The Filter stage applies multi-criteria selection based on textual quality, content relevance, and difficulty distribution. Finally, the Verify stage conducts manual validation to ensure problem readability, completeness, and test case accuracy. This end-to-end pipeline transforms raw datasets into a high-quality, standardized corpus suitable for reinforcement learning training.

\subsubsection{Processing}
\label{Processing}

The processing stage implements standardization through five core steps. Translation converts non-English problems (e.g., Japanese AIZU problems) to English for uniform accessibility. Noise removal addresses multiple data quality problems: missing images that affect problem comprehension, incomplete mathematical formulas and symbols, incomplete tables or graphics, irrelevant collected content including links and advertisements, incomplete problem descriptions, and content quality concerns. Problems that cannot be adequately processed are automatically removed while preserving original problem accuracy without subjective modifications. Test case optimization tackles two challenges. For problems either lacking test cases but containing reference solutions, or having test cases with noise from web, LLM generates comprehensive test cases. While LLM cannot solve every problem, it excels at generating test case inputs and considering boundary conditions. By executing reference solution code with these inputs, the system obtains guaranteed accurate outputs, enabling automatic generation of comprehensive, accurate, and appropriately-sized test suites. For problems with excessive test cases (hundreds per problem, resulting in datasets exceeding 100GB across thousands of problems), the framework selects the 15 longest test cases under the assumption that length correlates with difficulty. This massive data volume influences processing, loading, training speed, and stability. The system filters out problems without test cases and those requiring functional validation where multiple correct answer formats exist. This processing stage inherently incorporates filtering mechanisms for both problem quality and test case selection.

Format standardization unifies prompt structures, with CF retaining its native format while other sources adopt LiveCodeBench format. This addresses the execution logic differences between LeetCode-style function completion and OJ-style input/output problems. Since both problem types are mixed together during model training, many original datasets lack clear format instructions, potentially causing models to solve problems correctly but use incorrect code formats.

\begin{figure*}[t!]
    \centering
    \includegraphics[width=0.99\textwidth]{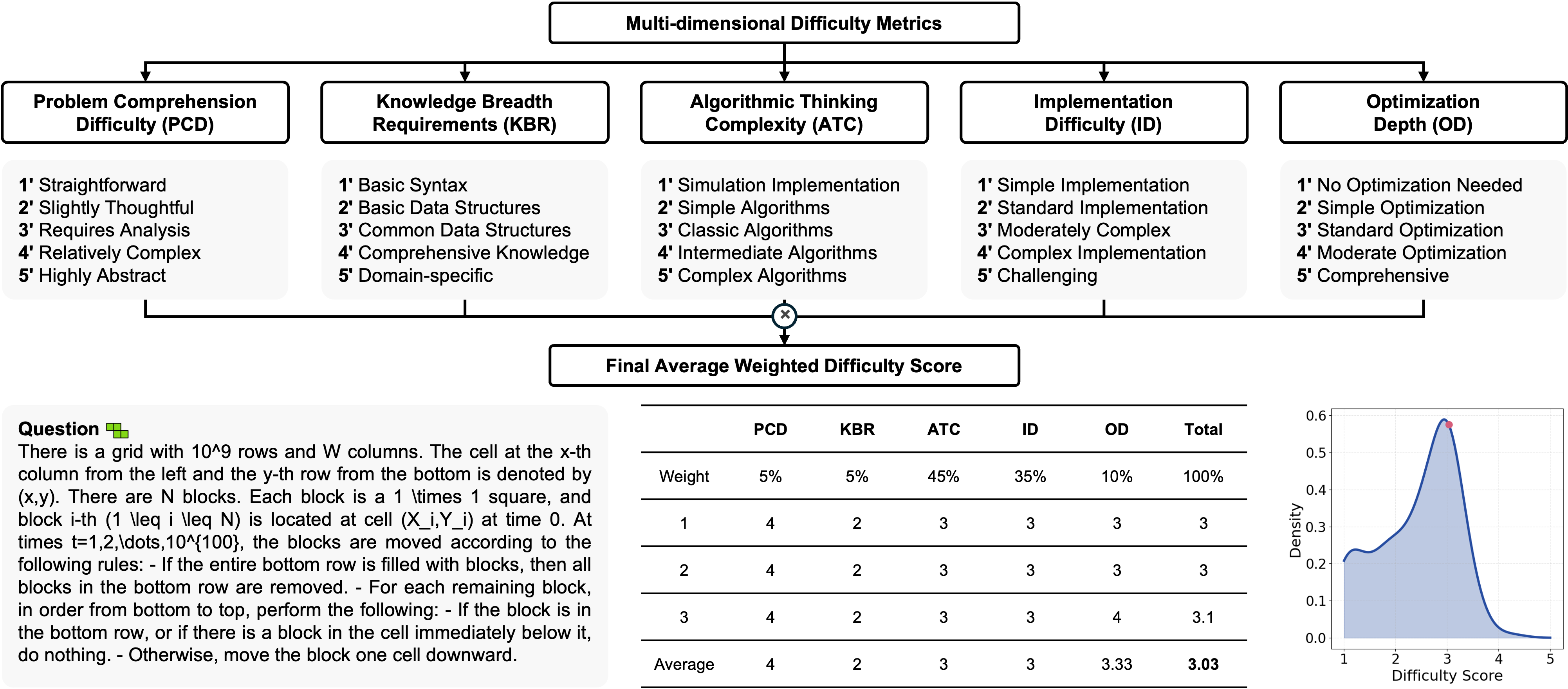}
    \caption{\textbf{Multi-dimensional Difficulty Metrics Framework.} \textit{Top:} Five-dimensional assessment system where each dimension is evaluated on a 1-5 scale with specific descriptors defining evaluation criteria. \textit{Bottom Left:} Example of scoring a question with three independent GPT-4O assessments. Each assessment rates all dimensions (for example, Assessment 1: PCD=4, KBR=2, ATC=3, ID=3, OD=3) and applies weights (ATC 45\%, ID 35\%, OD 10\%, PCD 5\%, KBR 5\%) to yield individual scores (3, 3, 3.1), which average to a final difficulty score of 3.03. \textit{Bottom Right:} Distribution curve showing that the difficulty score of the example problem (3.03) positions it in the top 30\% of the dataset, indicating a relatively challenging problem.}
    \label{Multi-dimensional Difficulty Metrics Framework}
\end{figure*}

\subsubsection{Filtering}
\label{Filtering}

The filtering stage applies multi-level criteria through hard requirements and adaptive selection mechanisms. Hard requirements require text-only problems, uniqueness through overlap identification, and unseen status relative to test sets. Adaptive requirements leverage quality improvements from the processing stage and implement difficulty-based selection tailored to specific datasets and model capabilities, utilizing a multidimensional difficulty matrix powered by LLM. Detailed description and application of this matrix are presented in Section \ref{Multi-dimensional Difficulty Metrics} and \ref{Case Study}.

Train-test separation employs 16-gram similarity analysis with a 0.22 threshold. Validation using AtCoder problems (the most recent training data sharing sources with LiveCodeBench) against LiveCodeBench v6 test set reveals that approximately 3\% of training data exceeds the 0.22 similarity threshold, yet no problems are identical to test set problems. This demonstrates the comprehensiveness and efficiency of the 16-gram 0.22 standard, which is subsequently applied across all datasets to ensure train-test separation.

\subsection{Automatic Difficulty Filtering}
\label{Automatic Difficulty Filtering}

\subsubsection{Predict-Calibrate-Select}
\label{Predict-Calibrate-Select}
Automatic Difficulty Filtering employs a three-stage predict-calibrate-select framework. Predict utilizes LLM to assess problem complexity through a multidimensional difficulty matrix, averaging three independent assessments per problem. Calibrate bridges predicted difficulty scores with actual model performance to establish difficulty category boundaries, where ground truth difficulty is measured as the success rate across four model attempts on each problem. Select applies the calibrated boundaries to filter out overly simplistic problems, retaining only those within target difficulty ranges appropriate for training objectives.

\begin{figure*}[t!]
    \centering
    \includegraphics[width=0.99\textwidth]{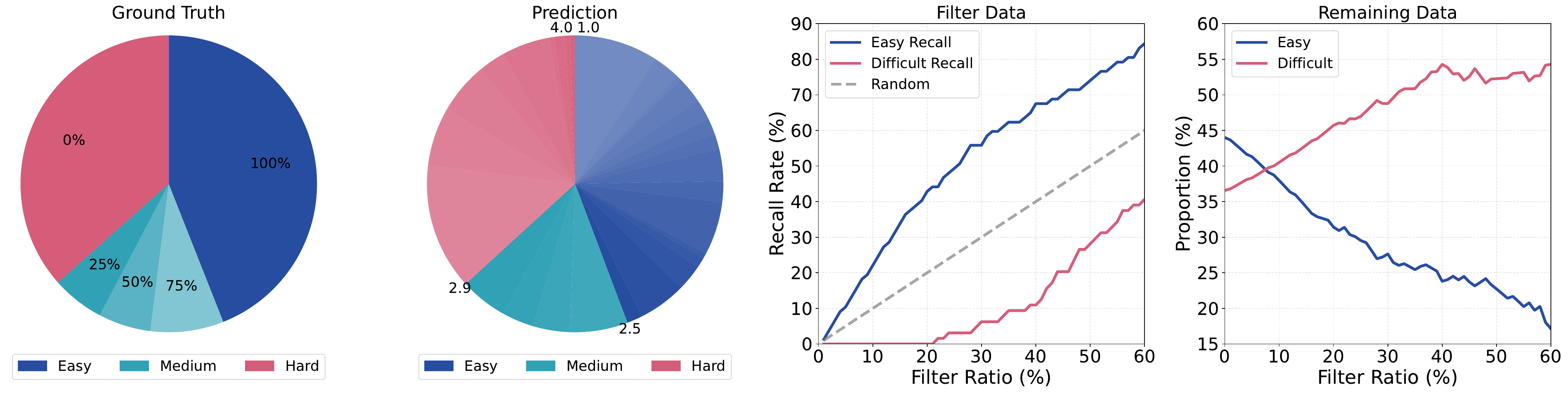}

    \caption{\textbf{Difficulty Filtering Case Study on LiveCodeBench v6.} \textit{Left:} Actual difficulty distribution based on model pass rates. \textit{Middle Left:} GPT-4O predicted difficulty calibrated against ground truth to determine boundaries at 2.5 (easy/medium) and 2.75 (medium/hard), achieving near-perfect distribution with actual distribution and enabling accurate difficulty-based filtering for subsequent data. \textit{Middle Right:} Recall curves showing the proportion of each difficulty category removed, where high easy recall (more than 60\% at 30\% filtering) versus low difficult recall demonstrates selective removal of simple problems. \textit{Right:} Difficulty composition of retained problems, where filtering 30\% of data changes the distribution from 45\% easy to under 25\% easy while increasing difficult problem proportion to over 50\%, confirming effective difficulty-based selection.}
    \label{Difficulty Filtering Case Study on LiveCodeBench v6}
\end{figure*}

\begin{figure*}[t!]
    \centering
    \includegraphics[width=0.99\textwidth]{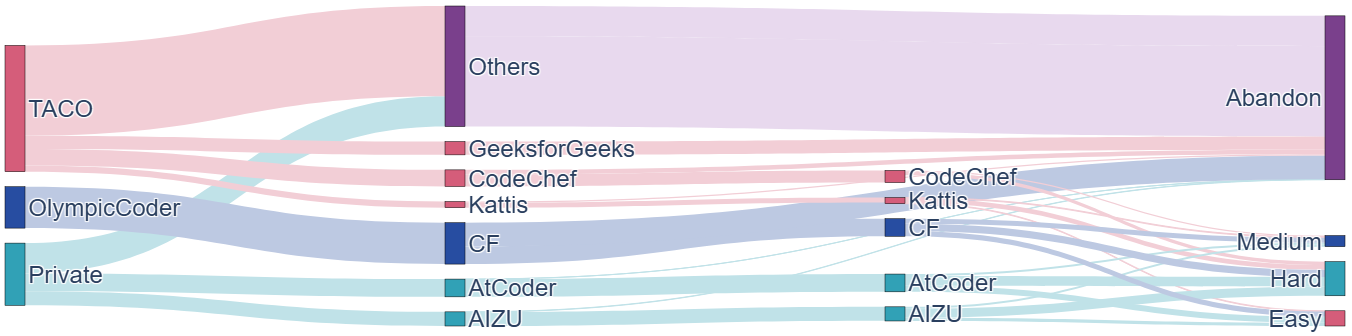}
    \caption{\textbf{Dataset Composition and Processing Flow.} Sankey diagram illustrating the complete data pipeline from diverse sources to final difficulty categorization. The dataset comprises exclusively real competitive programming problems without generated content. Open-source contributions undergo substantial filtering, with most problems abandoned and primarily difficult problems retained. Private collections contribute the majority of challenging and recent problems across platforms, resulting in a challenging and high-quality dataset.}
    \label{Dataset Composition and Processing Flow}
\end{figure*}

\subsubsection{Multi-dimensional Difficulty Metrics}
\label{Multi-dimensional Difficulty Metrics}

Multi-dimensional Difficulty Metrics (Figure \ref{Multi-dimensional Difficulty Metrics Framework}) decomposes problem complexity into five weighted dimensions. The difficulty matrix and its associated weights are designed drawing on cognition, evaluation, and software theories, including Bloom's Taxonomy of Educational Objectives \citep{bloom1956bloom}, General Evaluation Dimensions \citep{zhou2025generalscalesunlockai}, McCabe Complexity Theory \citep{mccabe1976complexity}, and Halstead Complexity Measures \citep{halstead1977elements}. Problem Comprehension Difficulty and Knowledge Breadth Requirements receive minimal weight as they primarily test semantic understanding and memory. Algorithmic Thinking Complexity and Implementation Difficulty receive substantial weight as they assess reasoning and programming capabilities. Each dimension uses a 1-5 scale with specific descriptors, enabling consistent evaluation across problems. The example demonstrates GPT-4O scoring a grid manipulation problem across three independent assessments, yielding an average difficulty of 3.08, placing it in the top 30\% of the dataset as a relatively challenging problem. The accuracy of this matrix is validated by the performance gains in Table \ref{results_table}.

\subsubsection{Case Study}
\label{Case Study}

Case Study demonstrates the complete predict-calibrate-select pipeline using LiveCodeBench (Figure \ref{Difficulty Filtering Case Study on LiveCodeBench v6}). First, GPT-4O evaluates problems using the multi-dimensional matrix, averaging three assessments per problem. Second, a subset undergoes empirical validation where Qwen-3-4B-thinking attempts each problem four times to establish ground truth difficulty through success rates. Calibration reveals optimal difficulty boundaries at 2.5 and 2.75 for distinguishing easy, medium, and hard categories, producing nearly identical predicted and empirical difficulty distributions. Finally, filtering removes all problems scoring below 2.5, eliminating 30\% of the total dataset while removing over 65\% of easy problems and preserving difficult problems. This process reduces easy problem ratio from approximately 40\% to under 20\% in the final dataset.

\begin{figure*}[t!]
    \centering
    \includegraphics[width=0.99\textwidth]{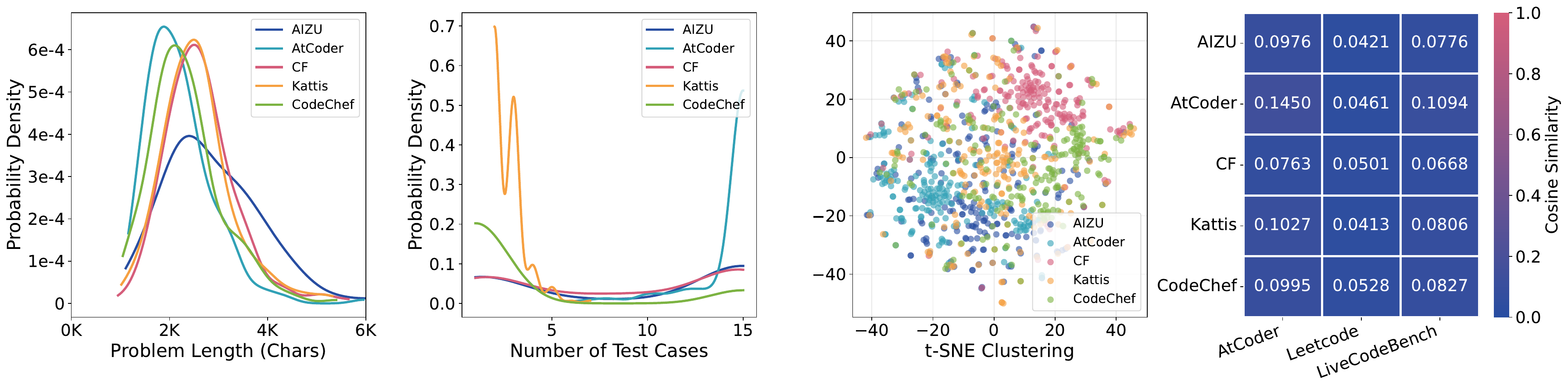}
    \caption{\textbf{Dataset Statistical Characteristics and Diversity Analysis.} \textit{Left:} Problem length distribution across platforms, with privately collected datasets (AIZU, AtCoder) showing comparable length profiles to open-source datasets (Kattis, CodeChef), where AIZU exhibits the longest average problem length. \textit{Middle Left:} Test case distribution where privately collected datasets exhibit more test cases per problem, with all datasets capped at 15 longest test cases following DeepCoder to prevent training pause from problems with hundreds of lengthy test cases. \textit{Middle Right:} t-SNE clustering visualization demonstrating clear platform separation and complementary coverage between privately collected and open-source datasets. \textit{Right:} Cosine similarity table where training datasets (y-axis) show low similarity (0.04-0.14) against test benchmarks (x-axis: AtCoder, LeetCode, LiveCodeBench), confirming no overlap between training and test sets.}
    \label{dataset_statistics}
\end{figure*}

\section{Dataset}
\label{Dataset}

\textbf{Dataset Composition.} The MicroCoder dataset comprises exclusively real competitive programming problems without generated data (Figure \ref{Dataset Composition and Processing Flow}). Open source contributions primarily derive from TACO and OlympicCoder datasets, while private data collection encompasses competition problems from diverse programming contest platforms. Rigorous quality and difficulty filtering reduces the initial corpus to 13,300 curated problems, with primary contributions from AIZU, AtCoder, CodeChef, and Kattis platforms.

\textbf{Dataset Statistics.} The MicroCoder dataset exhibits diverse statistical characteristics across platforms, as illustrated in Figure \ref{dataset_statistics}. Problem length distributions demonstrate consistency across sources, with most problems grouped between 1K-3K characters, while AIZU problems exhibit the longest average length. Test case distributions reveal an advantage of privately collected datasets, which contain more test cases per problem compared to open-source datasets. To prevent training pause caused by problems with hundreds of lengthy test cases, we cap all datasets at the 15 longest test cases, following DeepCoder's approach. The t-SNE clustering analysis demonstrates clear platform separation, confirming that privately collected and open-source datasets provide complementary coverage rather than redundant content. Furthermore, cosine similarity analysis between training datasets and test benchmarks (AtCoder, LeetCode, LiveCodeBench) shows consistently low similarity scores, from 0.04 to 0.14, validating rigorous train-test separation with zero overlap and ensuring reliable evaluation.

\section{Experimental Design}
\label{Experimental Design}

\textbf{Benchmarks}. We evaluate dataset quality on competitive programming benchmarks including AtCoder, LeetCode, from LiveCodeBench v6 \citep{jain2025livecodebench}, covering strictly unseen problems released after the model training cutoff. These benchmarks span multiple difficulty levels (easy, medium, hard) and provide comprehensive assessment of model performance on diverse algorithmic challenges. Evaluation employs the official LiveCodeBench v6 testing framework and code execution infrastructure to ensure standardized and reproducible results.

\textbf{Evaluation.} Model performance is measured by accuracy, where each problem receives a binary score: 1 if the generated solution passes all test cases, and 0 otherwise. We conduct four independent inference attempts per problem and report the average accuracy across these attempts. Results are presented across
difficulty categories and overall benchmarks.

\textbf{Baselines.} We compare MicroCoder against DeepCoder dataset \citep{deepcoder2025}, a widely-used and open-source competitive programming dataset. Both datasets are evaluated under identical training and inference configurations to ensure fair comparison.

\textbf{Models.} Experiments employ Qwen3-4B-Instruct-2507 \citep{yang2025qwen3technicalreport} trained from its official checkpoint. 

\textbf{Algorithms.} For reinforcement learning training, we utilize GRPO and DAPO \citep{yu2025dapo}. DAPO removes KL loss and employs high clipping to encourage the model to give more diverse solutions.

\textbf{Hyperparameters.} Training with GRPO and DAPO algorithms employs the following configurations: maximum response length of 8K tokens, temperature of 1.2, training batch size of 64, learning rate of 1e-6, 8 samples per query, and binary 0-1 accuracy as the reward.

\begin{figure*}[t!]
    \centering
    \includegraphics[width=0.99\textwidth]{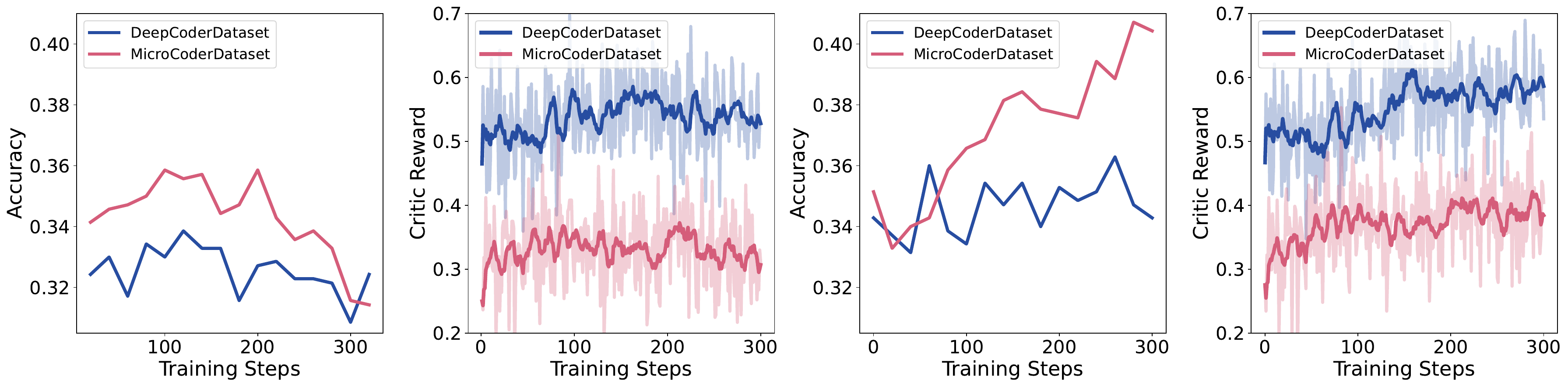}
    \caption{\textbf{Training Dynamics Comparison Between DeepCoder and MicroCoder Datasets.} Training curves under GRPO (left) and DAPO (right) algorithms, with accuracy on LiveCodeBench v6 test set and critic reward on training set. Darker lines represent smoothed curves over lighter raw training data. MicroCoder consistently achieves higher test accuracy, demonstrating stronger capability improvement for code generation models. DAPO, which removes KL loss and employs high clipping to encourage diverse solutions, yields better performance and more stable training than GRPO. MicroCoder exhibits lower training rewards than DeepCoder, indicating higher difficulty, yet achieves greater test improvement at comparable training progress, validating the effectiveness of challenging problems for model generalization.}
    \label{results_figures}
\end{figure*}

\begin{table*}[t!]
\centering
\fontsize{8pt}{9pt}\selectfont
\begin{tabular}{lccc>{\columncolor{gray!20}}c ccc>{\columncolor{gray!20}}c ccc>{\columncolor{gray!20}}c}
\hline
\multirow{2}{*}{} & \multicolumn{4}{c}{\textbf{AtCoder}} & \multicolumn{4}{c}{\textbf{LeetCode}} & \multicolumn{4}{c}{\textbf{LiveCodeBench}} \\
& Easy & Medium & Hard & \multicolumn{1}{>{\columncolor{gray!20}}c|}{All} & Easy & Medium & Hard & \multicolumn{1}{>{\columncolor{gray!20}}c|}{All} & Easy & Medium & Hard & All \\
\hline
\multicolumn{13}{c}{\textbf{Dataset Comparison}} \\
\multicolumn{13}{l}{\textbf{\textit{GRPO, Qwen3 4B, 8K}}} \\
APPS & 96.2 & 36.5 & 11.3 & 36.8 & 79.4 & 14.4 & 1.3 & 27.8 & 89.5 & 25.5 & 8.7 & 33.6 \\
DeepCoder & 96.2 & 40.4 & 10.4 & 37.3 & 73.5 & 19.2 & 0.0 & 27.8 & 87.2 & 29.8 & 7.8 & 33.9 \\
MicroCoder (Ours) & 97.1 & 47.1 & 11.3 & \textbf{39.5} & 76.5 & 18.3 & 3.7 & \textbf{29.4} & 89.0 & 32.7 & 9.4 & \textbf{35.9} \\
Delta & \textcolor{darkgreen}{+0.9} & \textcolor{darkgreen}{+6.7} & \textcolor{darkgreen}{+0.9} & \textcolor{darkgreen}{+2.2} & \textcolor{darkgreen}{+3.0} & \textcolor{darkred}{-0.9} & \textcolor{darkgreen}{+3.7} & \textcolor{darkgreen}{+1.6} & \textcolor{darkgreen}{+1.8} & \textcolor{darkgreen}{+2.9} & \textcolor{darkgreen}{+1.6} & \textcolor{darkgreen}{+2.0} \\
\multicolumn{13}{l}{\textbf{\textit{DAPO, Qwen3 4B, 8K}}} \\
APPS & 99.0 & 39.4 & 12.9 & 39.1 & 79.4 & 24.0 & 2.5 & 32.1 & 91.3 & 31.7 & 10.3 & 36.6 \\
CodeContests & 98.1 & 44.2 & 11.3 & 39.1 & 76.5 & 20.2 & 0 & 29.0 & 89.5 & 32.2 & 8.4 & 35.4 \\
KodCode RL & 99.0 & 46.2 & 11.3 & 39.7 & 77.9 & 26.0 & 2.5 & 32.5 & 90.7 & 36.1 & 9.1 & 37.1 \\
KodCode Instruct & 98.1 & 46.2 & 10.0 & 38.8 & 83.8 & 23.1 & 5.0 & 33.7 & 92.4 & 34.6 & 8.7 & 37.0 \\
DeepCoder & 98.1 & 40.4 & 12.1 & 38.6 & 77.9 & 24.0 & 3.7 & 32.1 & 90.1 & 32.2 & 10.0 & 36.3 \\
MicroCoder (Ours) & 99.0 & 49.0 & 14.6 & \textbf{42.2} & 83.8 & 33.7 & 5.0 & \textbf{38.1} & 93.0 & 41.3 & 12.2 & \textbf{40.7} \\
Delta & \textcolor{darkgreen}{+0} & \textcolor{darkgreen}{+2.8} & \textcolor{darkgreen}{+3.3} & \textcolor{darkgreen}{+2.5} & \textcolor{darkgreen}{+5.9} & \textcolor{darkgreen}{+7.7} & \textcolor{darkgreen}{+2.5} & \textcolor{darkgreen}{+5.6} & \textcolor{darkgreen}{+2.3} & \textcolor{darkgreen}{+5.2} & \textcolor{darkgreen}{+3.1} & \textcolor{darkgreen}{+3.6} \\
\hline
\multicolumn{13}{c}{\textbf{Model Ablation}} \\
\multicolumn{13}{l}{\textbf{\textit{DAPO, DS 8B, 16K}}} \\
APPS & 98.1 & 25.0 & 2.9 & 30.1 & 92.6 & 22.1 & 3.7 & 35.3 & 95.9 & 23.6 & 3.1 & 32.0 \\
CodeContests & 100 & 34.6 & 2.5 & 32.6 & 88.2 & 26.0 & 2.5 & 35.3 & 95.3 & 30.3 & 2.5 & 33.6 \\
KodCode RL & 100 & 21.2 & 2.5 & 29.5 & 91.2 & 26.0 & 1.3 & 35.7 & 96.5 & 23.6 & 2.2 & 31.7 \\
DeepCoder & 100 & 33.7 & 7.1 & 34.8 & 85.3 & 25.0 & 1.3 & 33.7 & 94.2 & 29.3 & 5.6 & 34.4 \\
MicroCoder (Ours) & 98.1 & 39.4 & 7.5 & \textbf{35.9} & 92.6 & 26.9 & 5.0 & \textbf{37.7} & 95.9 & 33.2 & 6.9 & \textbf{36.6} \\
Delta & \textcolor{darkred}{-1.9} & \textcolor{darkgreen}{+5.7} & \textcolor{darkgreen}{+0.4} & \textcolor{darkgreen}{+1.1} & \textcolor{darkgreen}{+7.3} & \textcolor{darkgreen}{+1.9} & \textcolor{darkgreen}{+3.7} & \textcolor{darkgreen}{+4.0} & \textcolor{darkgreen}{+1.7} & \textcolor{darkgreen}{+3.9} & \textcolor{darkgreen}{+1.3} & \textcolor{darkgreen}{+2.2} \\
\multicolumn{13}{l}{\textbf{\textit{DAPO, Qwen3 14B, 8K}}} \\
APPS & 100 & 50 & 13.8 & 42.2 & 70.6 & 20.2 & 5.0 & 29.0 & 88.4 & 35.1 & 11.6 & 37.4 \\
CodeContests & 98.1 & 50.0 & 11.3 & 40.4 & 85.3 & 18.3 & 2.5 & 31.3 & 93.0 & 34.1 & 9.1 & 37.1 \\
KodCode RL & 97.1 & 37.5 & 6.7 & 34.8 & 76.5 & 24.0 & 2.5 & 31.3 & 89.0 & 30.8 & 5.6 & 33.6 \\
DeepCoder & 99.0 & 51.9 & 13.3 & 42.2 & 83.8 & 26.9 & 5.0 & 35.3 & 93.0 & 39.4 & 11.3 & 39.7 \\
MicroCoder (Ours) & 99.0 & 50.0 & 17.5 & \textbf{44.0} & 91.2 & 31.7 & 5.0 & \textbf{39.3} & 95.9 & 40.9 & 14.4 & \textbf{42.3} \\
Delta & \textcolor{darkgreen}{+0} & \textcolor{darkred}{-1.9} & \textcolor{darkgreen}{+4.2} & \textcolor{darkgreen}{+1.8} & \textcolor{darkgreen}{+7.4} & \textcolor{darkgreen}{+4.8} & \textcolor{darkgreen}{+0} & \textcolor{darkgreen}{+4.0} & \textcolor{darkgreen}{+2.9} & \textcolor{darkgreen}{+1.5} & \textcolor{darkgreen}{+3.1} & \textcolor{darkgreen}{+2.6} \\
\hline
\multicolumn{13}{c}{\textbf{Component Ablation}} \\
\multicolumn{13}{l}{\textbf{\textit{GRPO, Qwen3 1.7B, 4K}}} \\
OlympicCoder Subset & 73.1 & 16.3 & 3.7 & 22.8 & 48.5 & 3.8 & 0 & 14.7 & 63.4 & 10.1 & 2.8 & 19.9 \\
DeepCoder Subset & 75.0 & 20.2 & 1.7 & 23.0 & 41.2 & 3.8 & 0 & 12.7 & 61.6 & 12.0 & 1.3 & 19.3 \\
MicroCoder Subset & 83.7 & 16.3 & 0.4 & \textbf{23.4} & 54.4 & 3.8 & 0 & \textbf{16.3} & 72.1 & 10.1 & 0.3 & \textbf{20.9} \\
\multicolumn{13}{l}{\textbf{\textit{DAPO, Qwen3 1.7B, 4K}}} \\
OlympicCoder Subset & 83.7 & 13.5 & 0 & 22.5 & 47.1 & 3.8 & 0 & 14.3 & 69.2 & 8.7 & 0 & 19.6 \\
DeepCoder Subset & 78.8 & 15.4 & 0.8 & 22.3 & 50.0 & 3.8 & 0 & 15.1 & 67.4 & 9.6 & 0.6 & 19.7 \\
MicroCoder Subset & 78.8 & 21.2 & 0 & \textbf{23.2} & 61.8 & 7.7 & 0 & \textbf{19.8} & 72.1 & 14.4 & 0 & \textbf{22.0} \\
\hline
\multicolumn{13}{c}{\textbf{Filtering Analysis}} \\
\multicolumn{13}{l}{\textbf{\textit{DAPO, Qwen3 8B, 8K}}} \\
OlympicCoder & 93.3 & 38.5 & 6.2 & 33.9 & 72.1 & 17.3 & 1.3 & 27.0 & 84.9 & 27.9 & 5.0 & 31.4 \\
OlympicCoder Filtered & 92.3 & 42.3 & 9.6 & \textbf{36.4} & 64.7 & 19.2 & 6.2 & \textbf{27.4} & 81.4 & 30.8 & 8.7 & \textbf{33.1} \\
KodCode & 88.5 & 40.4 & 5.0 & 32.6 & 69.1 & 13.5 & 1.3 & 24.6 & 80.8 & 26.9 & 4.1 & 29.7 \\
KodCode Filtered & 89.4 & 40.4 & 7.9 & \textbf{34.4} & 69.1 & 18.3 & 0 & \textbf{26.2} & 81.4 & 29.3 & 5.9 & \textbf{31.4} \\
\hline
\end{tabular}
\caption{\textbf{Performance Evaluation Across Model Scales and Dataset Sources.} Performance evaluation across different model scales using GRPO and its variants. Delta represents absolute improvement, Delta\% represents relative improvement, with green indicating gains, red indicating decrease, and bold indicating the best performance. MicroCoder consistently outperforms DeepCoder across all benchmarks and difficulty levels, achieving improvements on medium and hard problems, such as +3.7 points on LeetCode Hard under GRPO and +9.7 points (+40.4\%) on LeetCode Medium under DAPO, demonstrating that difficulty-based filtering achieves performance gains on challenging problems. Additional evaluations across model scales and data sources demonstrate that MicroCoder's advantages increase with larger models, and its advantage stems from incorporating recent, challenging problems rather than merely filtering existing data.}
\label{results_table}
\end{table*}

\section{Results}
\label{Results}

\subsection{Main Results Analysis}
\label{Main Results Analysis}

\textbf{Training Dynamics.} Figure \ref{results_figures} demonstrates consistent advantages of MicroCoder under both GRPO and DAPO algorithms. MicroCoder achieves higher test accuracy throughout training while exhibiting lower critic rewards on the training set, indicating higher problem difficulty that nevertheless translates to stronger model generalization. This difference between training rewards and test performance validates our hypothesis that challenging problems, despite being harder to solve during training, drive more effective capability improvement. DAPO yields higher performance on test sets and more stable training dynamics compared to the GRPO method, attributed to its removal of KL loss and high clipping method that encourages solution diversity, achieving greater gains on challenging problems.

\textbf{Overall Performance.} Table \ref{results_table} presents comprehensive benchmark results comparing DeepCoder and MicroCoder datasets under identical training configurations. MicroCoder consistently outperforms DeepCoder across all benchmarks and difficulty levels. Under GRPO training, MicroCoder achieves absolute improvements of +2.2 points on AtCoder, +1.6 points on LeetCode, and +2.0 points on LiveCodeBench overall accuracy. The advantages become more pronounced with DAPO, where MicroCoder delivers +3.6 points on AtCoder, +6.0 points on LeetCode, and +4.4 points on LiveCodeBench, representing relative improvements of 9.3\%, 18.7\%, and 12.1\% respectively.

\textbf{Difficulty-Specific Analysis.} Performance gains exhibit clear difficulty-dependent characteristics, validating the effectiveness of difficulty-based filtering. On easy problems, improvements remain relatively modest as both datasets approach performance convergence. Medium difficulty problems demonstrate substantial gains, with MicroCoder achieving +8.6 to +9.7 points under DAPO compared to DeepCoder across benchmarks, including a remarkable +40.4\% relative improvement on LeetCode Medium. Hard problems show the largest relative improvements, with +20.7\% to +22.0\% gains on LiveCodeBench and AtCoder Hard problems under DAPO. These results confirm that performance improvements primarily stem from medium and hard problems where model capabilities are most stretched, demonstrating that challenging problem selection leads to performance gains across difficulty levels.

\subsection{Ablation Studies}
\label{Ablation Studies}

To analyze the influencing variables and sources of performance gains, we conduct experiments across different model sizes and dataset sources using GRPO and its variants. Results reveal that MicroCoder consistently outperforms alternative sources across both scales. On 1.7B models under DAPO, MicroCoder achieves 23.2\% on AtCoder and 22.0\% on LiveCodeBench, substantially exceeding OlympicCoder (22.5\%, 19.6\%) and DeepCoder (22.3\%, 19.7\%). When scaling to 4B models, MicroCoder delivers 42.2\% on AtCoder and 40.7\% on LiveCodeBench, demonstrating larger absolute gains over baselines compared to smaller models.

These results reveal several insights. First, as model capabilities increase, MicroCoder's advantages become more pronounced, validating that challenging and recent problems contribute more when models have sufficient capacity to leverage them. Second, MicroCoder's advantage stems not merely from filtering existing problems, but from incorporating recent, challenging problems that improve model capabilities. Third, DAPO consistently delivers larger improvements than GRPO across all settings, demonstrating that diversity-driven training algorithms achieve greater gains on challenging problems. These findings confirm the importance of data recency and difficulty in training the latest code generation models.

\section{Conclusions}
\label{Conclusions}

We present a systematic approach to high-quality dataset creation through difficulty-aware curation. Our four-stage Data Processing Framework and Automatic Difficulty Filtering method address limitations in existing datasets, producing the MicroCoder dataset of 13,300 curated competitive programming problems. Evaluations on LiveCodeBench v6 demonstrate 3× larger performance gains compared to baseline datasets, with substantial improvements on medium and hard problems. These results validate that difficulty-aware curation leads to performance improvements, providing insights for dataset creation in code generation.

\textbf{Future Directions.} Promising directions include extending our framework to multiple programming languages, using additional difficulty dimensions for specific models, applying difficulty-aware curation to other code-related tasks such as program correction and code translation, and developing dynamic difficulty assessment mechanisms that adapt to model capabilities. Related work is helpful for understanding this project \citep{li-etal-2025-500xcompressor, li2025ptmoe, li2025flexilora, li-etal-2025-prompt, D4DD00307A, zhou2025generalscalesunlockai, li-etal-2025-reasongraph, li2025a}

\bibliography{custom}

\newpage

\appendix

\end{document}